\documentclass[conference,a4paper]{IEEEtran}
\IEEEoverridecommandlockouts

\usepackage{cite}
\usepackage{amsmath,amssymb,amsfonts}
\usepackage{graphicx}
\usepackage{textcomp}
\usepackage{xcolor}
\usepackage[most]{tcolorbox}
\usepackage{booktabs}
\usepackage{url}

\usepackage[caption=false,font=footnotesize]{subfig}

\begin{document}

\title{Quantifying Affective Bias in Low-Resource Media: Large-Scale Emotion Profiling of Bengali Headlines}

\author{
\IEEEauthorblockN{
Mohd Ruhul Ameen\IEEEauthorrefmark{1},
Akif Islam\IEEEauthorrefmark{1},
Ayesha Siddiqua\IEEEauthorrefmark{2},
Abu Saleh Musa Miah\IEEEauthorrefmark{3},
Jungpil Shin\IEEEauthorrefmark{3}}
\IEEEauthorblockA{ameensunny242@gmail.com, iamakifislam@gmail.com,
ayesha.siddiqua@nub.ac.bd, musa@u-aizu.ac.jp, jpshin@u-aizu.ac.jp}
\IEEEauthorblockA{\IEEEauthorrefmark{1}Department of Computer Science and Engineering, University of Rajshahi, Rajshahi 6205, Bangladesh}
\IEEEauthorblockA{\IEEEauthorrefmark{2}Department of Computer Science and Engineering, Northern University Bangladesh, Dhaka, Bangladesh}
\IEEEauthorblockA{\IEEEauthorrefmark{3}School of Computer Science and Engineering, University of Aizu, Aizuwakamatsu 965-8580, Japan}
}

\maketitle

\begin{abstract}
News media can influence readers not only through the events they report but also through the emotional tone used to present them. This issue is especially important in digital news environments, where headlines often shape first impressions before readers open the full article. This study examines affective framing in Bengali digital journalism through corpus level emotion analysis of news headlines. Using zero shot inference with Gemma 3 4B, we analyzed 300{,}000 Bengali news headlines to estimate the dominant emotion and overall affective tone of each headline. The results show that negative emotion labels, particularly anger, sadness, disappointment, and fear, appear frequently in the analyzed corpus. A small pilot validation on 200 manually reviewed headlines suggests that the model can provide useful emotion estimates, although the results should be interpreted as computational estimates rather than a complete benchmark. Based on these findings, we propose a conceptual bias sensitive news interface that visualizes emotional cues across news sources and helps readers notice affective framing patterns in daily news.
\end{abstract}


\section{Introduction}\label{sec:intro}

News media do more than report events. They also shape how readers understand and emotionally respond to those events. The choice of words, emphasis, and tone in a news story can influence public interpretation, empathy, and sharing behavior \cite{hasell2020partisan,baum2008new,lazer2018science}. The same event may therefore appear calm in one outlet and alarming in another, showing that news often contains an affective layer beyond factual reporting.

This affective layer is especially important in digital news environments, where platforms reward attention and engagement. Prior work has shown that negative and high arousal content, such as anger, fear, and outrage, can attract greater attention and circulate more widely than neutral content \cite{soroka2015bad}. Such patterns may encourage emotionally charged framing and contribute to affective polarization, where public reactions are shaped strongly by emotion \cite{iyengar2019origins,boxell2022cross}.

Headlines play a central role in this process because they are often the first and sometimes the only part of an article that readers see. Many users react to headlines on social media and news feeds without reading the full article, which means headlines can strongly shape first impressions \cite{gabielkov2016social,ecker2014effects,andrew2007media}. This issue becomes more complex in multilingual and regional media ecosystems, where emotional expression depends on language, culture, and local context \cite{mohammad2016semeval}.

This study examines Bengali digital news, a large media ecosystem that remains underexplored in affective computing and media analysis. We analyze 300{,}000 Bengali news headlines using Gemma 3 4B to estimate dominant emotion and overall affective tone. Rather than proposing a new emotion classification algorithm, this work presents an empirical and design oriented study of affective framing in a low resource language context. The findings are interpreted as computational estimates, not evidence of editorial intent. Building on these observations, we propose a conceptual bias sensitive news interface that visualizes emotional cues across outlets and supports more transparent news consumption.

\section{Related Work}\label{sec:related}

\subsection{From Sentiment to Fine Grained Emotion}
Early work on emotion and sentiment analysis relied heavily on lexicons and rule based resources \cite{mohammad2013nrc}. More recent studies have moved toward transformer based models, including BERT and related architectures, which have improved the ability to recognize emotions from text \cite{demszky2020goemotions,sosea2021emotion}. However, much of the work in news and social media analysis still focuses on coarse sentiment categories such as positive, negative, and neutral. This representation is useful for broad polarity analysis, but it does not fully capture the difference between emotions such as anger, fear, sadness, and disappointment. These emotions may share a negative tone, but they can influence readers in different ways. Fine grained taxonomies such as GoEmotions provide a more detailed view of emotional expression \cite{demszky2020goemotions}, yet their use in digital journalism remains limited.

\subsection{Affective Framing in News Media}
News bias research has traditionally focused on political slant, source credibility, lexical framing, and misinformation \cite{fan2019plain,baly2020we}. These directions are important, but they often give less attention to the emotional tone through which news is presented. Headlines are especially important because they are designed to summarize, attract attention, and guide interpretation before a reader opens the full article. Prior media studies suggest that negative news often receives stronger attention than neutral or positive news \cite{soroka2015bad}. This does not necessarily imply intentional manipulation by media outlets. Rather, it indicates that emotional framing can become part of the structure of digital news production and consumption. Existing emotion models are often trained on social media or conversational datasets, which may not fully represent the more formal and compressed language of journalistic headlines \cite{strapparava2007semeval}. Therefore, affective framing in news remains an important area for further study.

\subsection{Low Resource and Multilingual Challenges}
Emotion analysis becomes more challenging in multilingual settings because emotional expression is shaped by language, culture, and social context \cite{mohammad2016semeval}. Multilingual models such as XLM R and mDeBERTa have improved cross lingual text classification and zero shot transfer \cite{conneau2020unsupervised,perez2021pysentimiento}. However, research attention remains uneven across languages. High resource languages have received extensive study, while Bengali remains comparatively underexplored in fine grained emotion analysis. Existing Bengali NLP studies have often focused on sentiment classification in social media comments or general text classification tasks\cite{akif_peft}. In contrast, Bengali digital journalism represents a large and socially important text domain that has not been sufficiently examined from an affective computing perspective.

\subsection{Research Gap and Contribution}
The reviewed literature shows three main gaps. First, most news analysis systems still focus on polarity rather than fine grained emotional framing. As a result, they may fail to distinguish between different negative emotions such as fear, anger, and sadness. Second, Bengali news remains underrepresented in large scale emotion analysis, despite the size and social importance of the Bengali language community. Third, current news aggregation systems mainly prioritize relevance, recency, and popularity, while offering limited support for emotional transparency. Related work on user-facing intermediaries for social media has proposed inline integrity and bias indicators as a way to surface such signals without restricting access \cite{ameen2025humanized}.

This work addresses these gaps through an empirical and design oriented study of Bengali news headlines. We apply zero shot inference with Gemma 3 4B to analyze 300{,}000 headlines and estimate their dominant emotional tone. The study does not claim to introduce a new classification algorithm. Instead, it contributes a large scale computational analysis of affective framing in a low resource language context and proposes a conceptual interface for making such emotional patterns more visible to readers.

\section{Methodology}\label{sec:methods}

\subsection{Data Collection and Preprocessing}

We collected a corpus of \textbf{824,784} Bengali news articles from major online news outlets, including \textit{Prothom Alo}, \textit{BDNews24}, \textit{Ittefaq}, and \textit{Samakal}. Each record contained the headline, article text, source name, and available publication metadata. Since processing the full corpus was computationally expensive, we selected \textbf{300,000} headlines for emotion analysis. The subset was sampled to preserve coverage across news sources, topic categories, and publication periods as much as possible.

\begin{table}[!t]
\caption{Sample Dataset of News Headlines and Content (Bangla)}
\label{tab:sample_dataset}
\centering
\includegraphics[width=\linewidth]{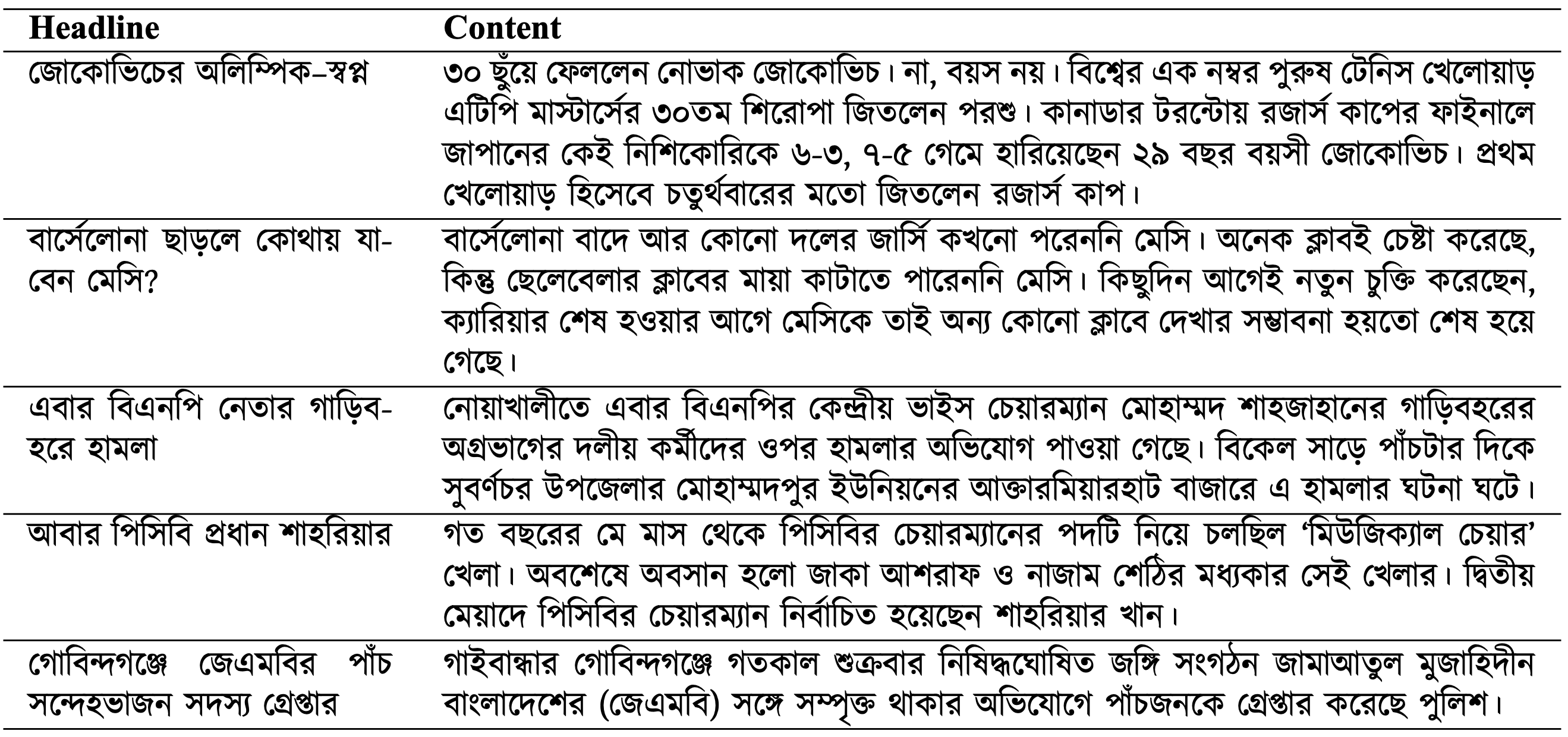}
\end{table}

The preprocessing stage focused on making the headline text suitable for model based analysis. We removed non Bengali entries, normalized the text encoding to UTF 8, and cleaned HTML tags, advertisements, duplicate spacing, and formatting noise. Very short or incomplete headlines were excluded because they often lacked enough context for emotion classification. The cleaned records were then linked with source and timestamp metadata to support later analysis of outlet level and temporal patterns.

\begin{table}[!t]
\caption{Top 10 Topics by Frequency in the Dataset}
\label{tab:topics_by_outlet}
\centering
\includegraphics[width=0.65\linewidth]{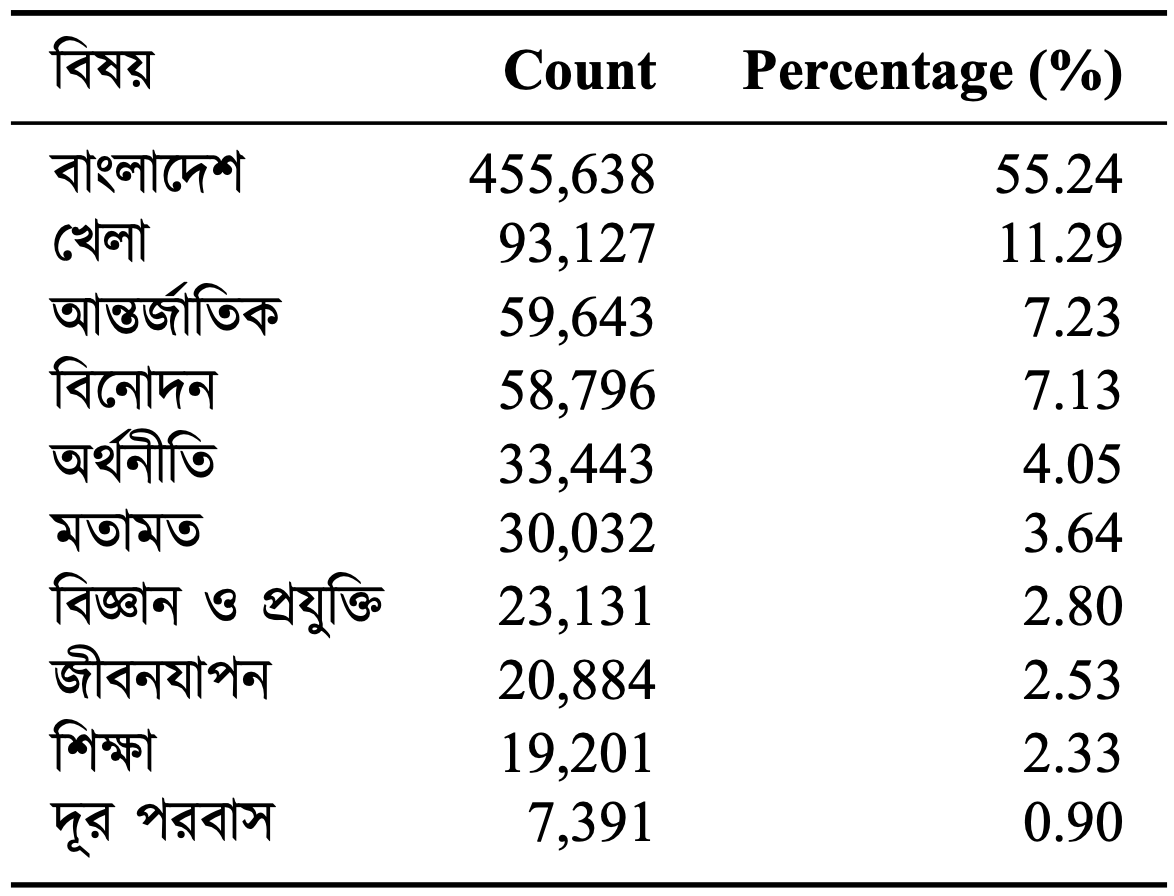}
\end{table}

\section{Emotion Classification Framework}\label{sec:method}

\subsection{Model Selection and Emotion Taxonomy}

We used \textbf{Gemma 3 4B} \cite{team2024gemma} for zero shot emotion classification. The model was selected because it can follow structured instructions and has shown useful performance in recent Bengali and low resource language tasks \cite{akif_peft}. This study does not propose a new emotion classification model. Instead, the goal is to examine whether an instruction following LLM can be used to support large scale affective analysis of Bengali news headlines.

The emotion labels were based on the \textbf{GoEmotions 28 taxonomy} \cite{demszky2020goemotions}. This taxonomy provides a more detailed view of emotion than simple positive, negative, and neutral sentiment categories. For each headline, the model was asked to identify the dominant emotion and return a structured JSON output. The use of JSON formatting helped reduce irregular responses and made the output easier to parse during large scale processing.

The following instruction style was used for zero shot inference:

\begin{tcolorbox}[
colback=gray!5,
colframe=black!70,
title=Prompt Template,
fonttitle=\bfseries,
boxrule=0.6pt,
arc=2mm
]
\small
You are an emotion classification assistant. Given a Bengali news headline, identify the dominant emotion from the GoEmotions label set. Return only a valid JSON object with the fields ``dominant\_emotion'', ``coarse\_tone'', and ``confidence''. Do not explain the answer.
\end{tcolorbox}

The output format was constrained as follows:

\begin{tcolorbox}[
colback=gray!5,
colframe=black!70,
title=Example Output Format,
fonttitle=\bfseries,
boxrule=0.6pt,
arc=2mm
]
\small
\texttt{\{"dominant\_emotion": "anger", "coarse\_tone": "negative",\\ "confidence": 0.82\}}
\end{tcolorbox}

\subsection{Inference Setup}

All inference was conducted locally on a single NVIDIA RTX 4090 GPU with 24 GB VRAM. To reduce processing time, three local Ollama endpoints were launched in parallel. Incoming requests were assigned to the endpoints using a round robin strategy. If one endpoint failed to return a valid response, the request was retried through another available endpoint.

\begin{table}[!t]
\caption{Endpoint utilization during inference. Each of the three local Ollama endpoints processed approximately one third of the total 300{,}000 news items. Two records failed to infer due to invalid Unicode character processing.}
\label{tab:endpoint_utilization}
\centering
\begin{tabular}{lrr}
\hline
\textbf{Endpoint} & \textbf{Number of Calls} & \textbf{Percentage (\%)} \\
\hline
\texttt{http://localhost:11435} & 99{,}999 & 33.33 \\
\texttt{http://localhost:11436} & 99{,}999 & 33.33 \\
\texttt{http://localhost:11437} & 100{,}000 & 33.34 \\
\hline
\textbf{Total Processed} & \textbf{300{,}000} & \textbf{100.00} \\
\hline
\end{tabular}
\end{table}

This setup reduced the average inference time from 0.70 seconds to 0.48 seconds per headline. The full 300{,}000 headline subset was processed in approximately 40 hours. As shown in Table~\ref{tab:endpoint_utilization}, the requests were distributed almost evenly across the three endpoints. Two records could not be processed because of invalid Unicode characters and were excluded from the final analysis.

\section{Results and Discussion}\label{sec:results}

\begin{table}[!t]
\caption{Pilot validation of Gemma 3 zero shot emotion classification on Bengali headlines (n=200)}
\label{tab:validation}
\centering
\footnotesize
\begin{tabular}{lrrr}
\hline
\textbf{Emotion} & \textbf{Precision} & \textbf{Recall} & \textbf{F1 Score} \\
\hline
Anger & 0.78 & 0.82 & 0.80 \\
Fear & 0.72 & 0.68 & 0.70 \\
Joy & 0.85 & 0.88 & 0.86 \\
Sadness & 0.76 & 0.74 & 0.75 \\
\hline
\textbf{Macro Avg} & \textbf{0.75} & \textbf{0.76} & \textbf{0.75} \\
\hline
\end{tabular}
\end{table}

Before analyzing the full dataset, we conducted a small pilot validation on 200 manually reviewed Bengali headlines. As shown in Table~\ref{tab:validation}, Gemma 3 achieved a macro average F1 score of 0.75 across the selected emotion categories. This result suggests that the model can provide useful emotion estimates for Bengali headlines. However, the validation set is limited in size and should be interpreted as a feasibility check rather than a complete benchmark.

The emotion classification results for 300{,}000 Bengali news headlines show a clear concentration of negative emotional labels. As shown in Table~\ref{tab:dist}, \textit{anger}, \textit{sadness}, \textit{disappointment}, and \textit{fear} together account for 50.42\% of the analyzed headlines. In comparison, \textit{neutral} accounts for 13.13\%, while \textit{joy} accounts for 8.72\%. These findings suggest that negative emotional framing is highly visible in the sampled Bengali headline corpus. Since the analysis is based on model inference, the results should be understood as estimated affective patterns rather than direct evidence of editorial intention.

\begin{table}[!b]
\centering
\caption{Dominant Emotion Distribution ($N=300{,}000$)}
\label{tab:dist}
\footnotesize
\begin{tabular}{lrr}
\toprule
\textbf{Emotion} & \textbf{Count} & \textbf{Percentage (\%)} \\
\midrule
Anger            & 51{,}806 & 17.269 \\
Neutral          & 39{,}399 & 13.133 \\
Sadness          & 35{,}372 & 11.791 \\
Disappointment   & 35{,}157 & 11.719 \\
Surprise         & 31{,}246 & 10.415 \\
Fear             & 28{,}926 & 9.642 \\
Joy              & 26{,}174 & 8.725 \\
Disapproval      & 11{,}919 & 3.973 \\
Annoyance        & 10{,}577 & 3.526 \\
Approval         & 5{,}607  & 1.869 \\
Disgust          & 4{,}534  & 1.511 \\
Optimism         & 4{,}070  & 1.357 \\
Excitement       & 2{,}664  & 0.888 \\
Relief           & 2{,}527  & 0.842 \\
Admiration       & 2{,}335  & 0.778 \\
Curiosity        & 2{,}221  & 0.740 \\
Caring           & 1{,}676  & 0.559 \\
Gratitude        & 1{,}292  & 0.431 \\
Pride            & 1{,}013  & 0.338 \\
Amusement        & 353      & 0.118 \\
Nervousness      & 310      & 0.103 \\
Desire           & 278      & 0.093 \\
Love             & 248      & 0.083 \\
Confusion        & 240      & 0.080 \\
Remorse          & 35       & 0.012 \\
Embarrassment    & 16       & 0.005 \\
Grief            & 5        & 0.002 \\
\bottomrule
\end{tabular}
\end{table}

\begin{figure*}[t]
    \centering
    \includegraphics[width=0.8\textwidth]{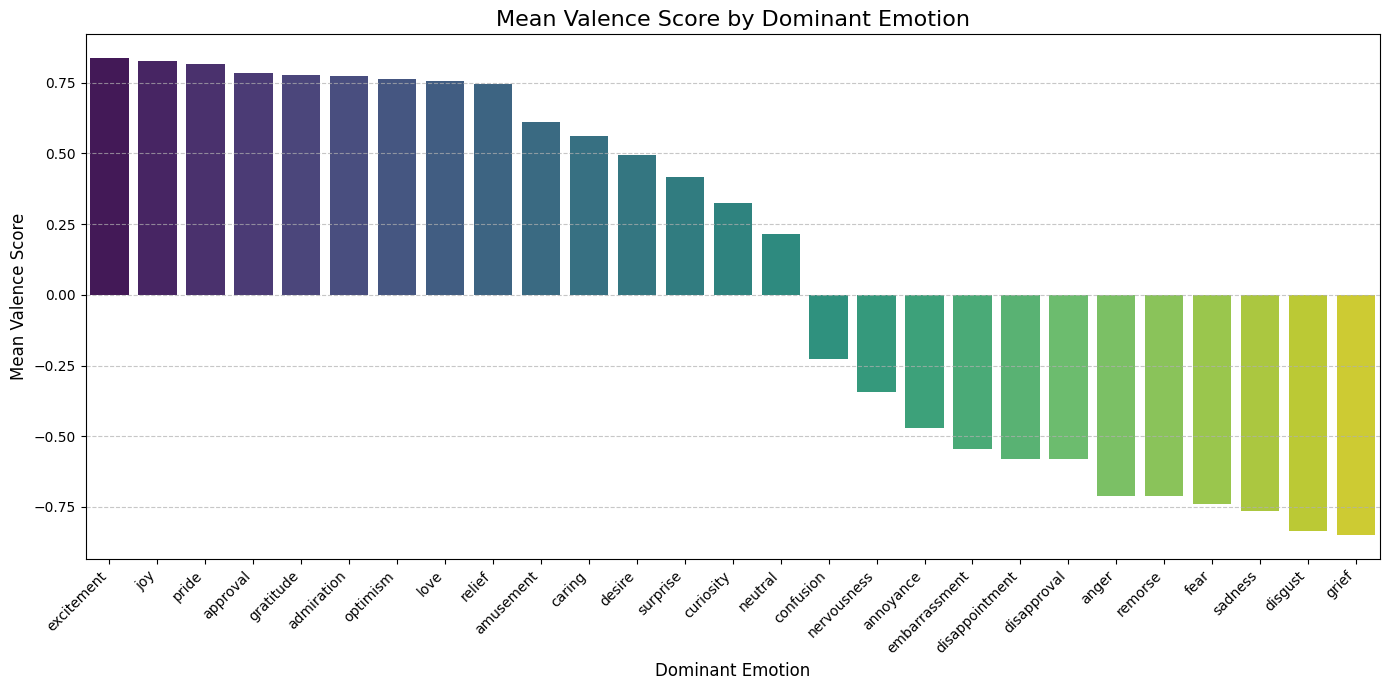}
    \caption{Mean valence score by dominant emotion. Positive emotions such as \textit{joy}, \textit{pride}, and \textit{gratitude} show higher valence, while emotions such as \textit{anger}, \textit{fear}, and \textit{sadness} show lower valence.}
    \label{fig:valence_score}
\end{figure*}

The use of the GoEmotions 28 taxonomy provides more detail than a simple positive, negative, and neutral sentiment division. For example, \textit{disappointment} appears in 11.72\% of the headlines, which is close to the proportion of \textit{sadness} at 11.79\%. Although both emotions are negative, they may reflect different kinds of framing. Sadness often suggests loss or suffering, while disappointment may point to unmet expectations, failure, or dissatisfaction. This distinction would be difficult to observe with binary sentiment analysis alone.

Some emotions appear very rarely in the corpus. Labels such as \textit{grief}, \textit{remorse}, and \textit{embarrassment} account for less than 0.02\% of the headlines. This may reflect the concise and public facing style of news headlines, where personal or introspective emotions are less common. However, these low frequency results should be interpreted carefully because rare classes are more difficult to validate with a small pilot sample.

A source specific analysis of \textit{Prothom Alo} ($n=197,000$) shows a similar pattern. The estimated negative emotion ratio is 51.82\%, with \textit{anger} at 18.77\% and \textit{fear} at 10.19\%. This result is consistent with earlier work on negativity bias in news consumption \cite{soroka2015bad}. However, the present study does not claim that outlets intentionally use negative emotion to influence readers. Rather, it shows that negative affective labels are common in the sampled headline data and that such patterns can be measured computationally.

\begin{figure}[!t]
    \centering
    \includegraphics[width=0.48\textwidth]{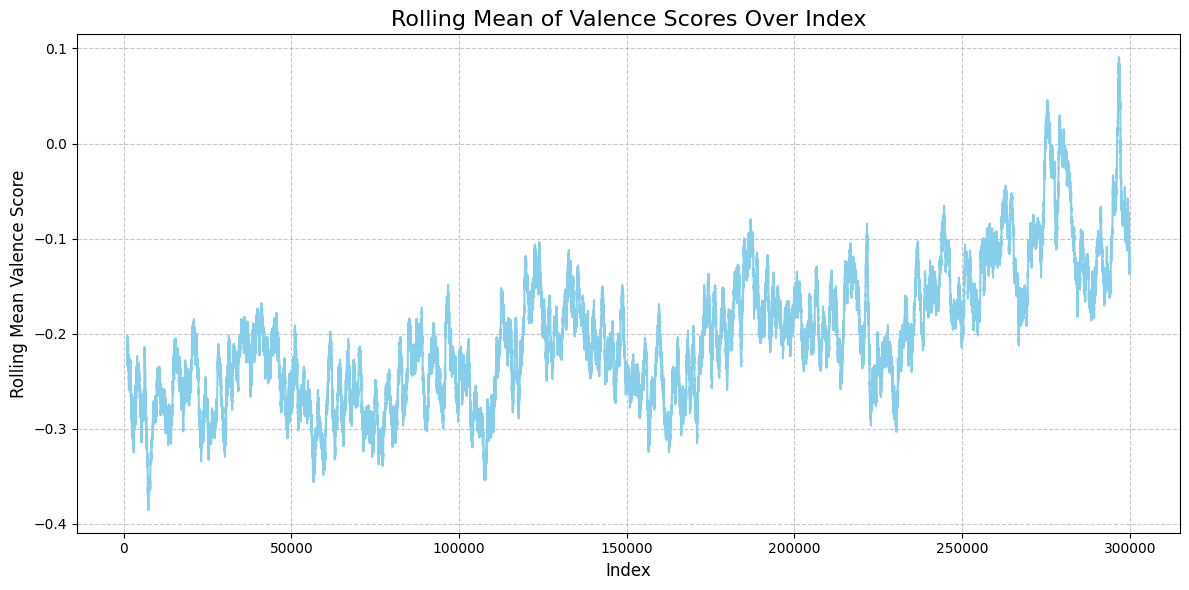}
    \caption{Rolling mean of valence scores over the headline index. The trend suggests changes in estimated emotional tone across the analyzed sample.}
    \label{fig:rolling_valence}
\end{figure}

\subsection{Computational Efficiency}

The local inference setup made it possible to process the selected corpus within a practical time frame. By running three Gemma 3 4B instances on a single NVIDIA RTX 4090 GPU, the average inference time was reduced from 0.70 seconds to 0.48 seconds per headline. This corresponds to an approximate 1.5$\times$ speed improvement. The full 300{,}000 headline subset was processed in about 40 hours, with an average throughput of approximately 125 items per minute. These results show that local LLM based emotion analysis can be feasible for medium scale news analysis when parallel inference endpoints are used.

\subsection{Implications for Media Design}

The findings suggest that emotion aware interfaces may help readers better understand the affective tone of news before engaging with full articles. Instead of treating news ranking only as a matter of relevance or recency, a news interface can also show whether a headline is estimated to contain anger, fear, sadness, joy, or neutral framing. Such information may support more reflective news consumption. At the same time, these design implications should be treated as hypotheses. This study does not include a user study, and therefore it cannot claim that the proposed interface improves reader behavior or reduces psychological distress.

\section{Proposed Design}\label{sec:design}

\begin{figure*}[!t]
\centering
\includegraphics[width=0.75\textwidth]{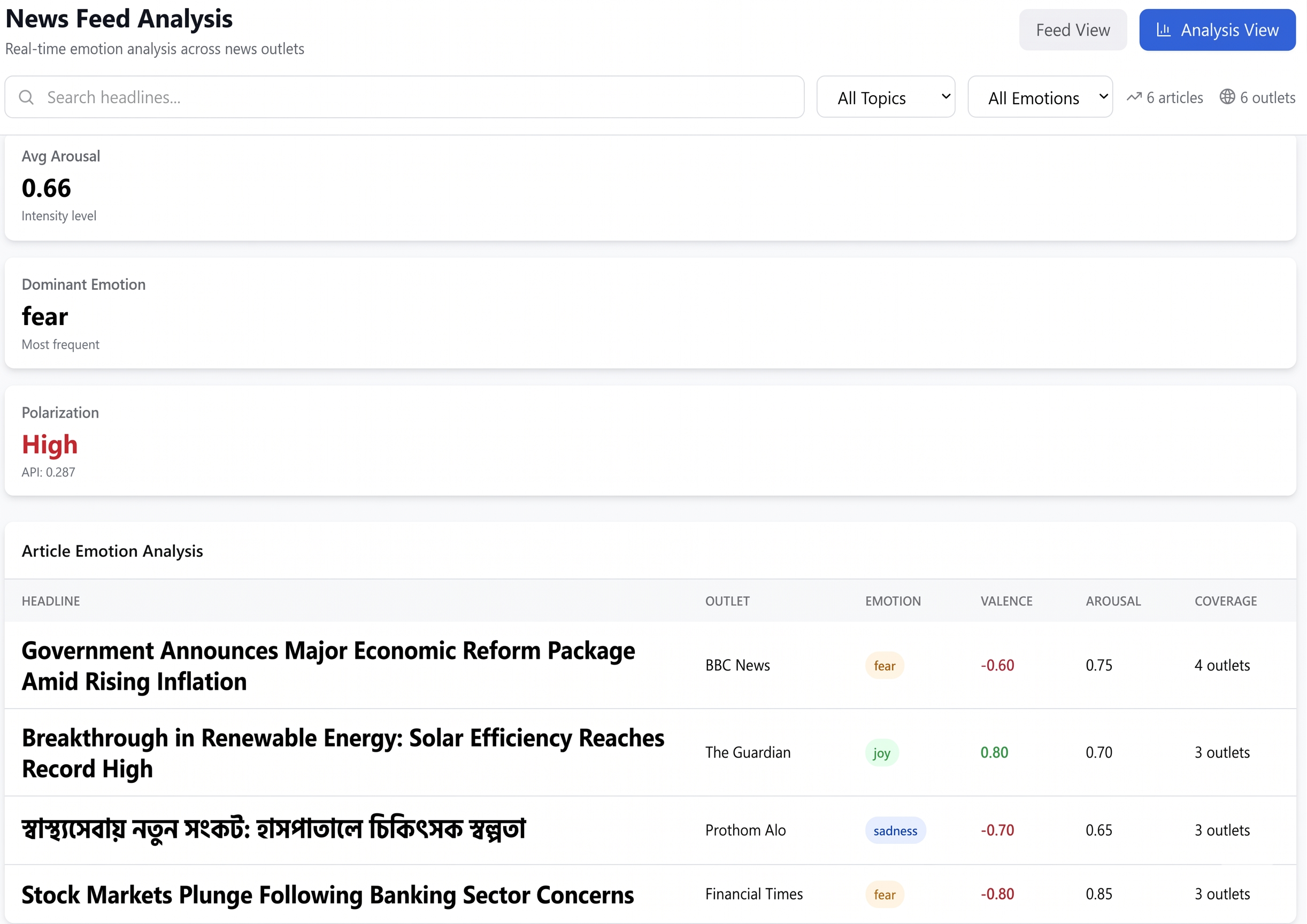}
\caption{News Feed Analysis dashboard displaying estimated valence, arousal, and dominant emotion patterns across outlets.}
\label{fig:news_feed}
\end{figure*}

We propose the \textbf{Bias Sensitive News Interface} as a conceptual design based on the emotion analysis results. The purpose of the interface is to show how affective information could be presented to readers in a news aggregation environment. The design is not evaluated with users in this version of the study, so it should be understood as a design proposal rather than a deployed system.

\subsection{System Overview}

The proposed system follows a simple visualization pipeline. News headlines are first processed by the emotion classification model. For each headline, the system stores the estimated dominant emotion, valence, and arousal. These values are then aggregated by source, topic, and time period. The interface uses these aggregated values to help users compare emotional patterns across news outlets and news cycles.

\subsection{News Feed Analysis View}

The \textit{News Feed Analysis} view in Fig.~\ref{fig:news_feed} presents a summary of the emotional tone of current headlines. It lists headlines with their estimated valence and arousal scores and allows users to compare how different outlets present similar topics. For example, one outlet may present an economic issue with a more negative tone, while another may frame the same issue more neutrally. The goal of this view is not to judge the correctness of a headline, but to make emotional framing easier to notice.

\subsection{Bias Sensitive News Interface}

\begin{figure}[!t]
\centering
\includegraphics[width=0.9\linewidth]{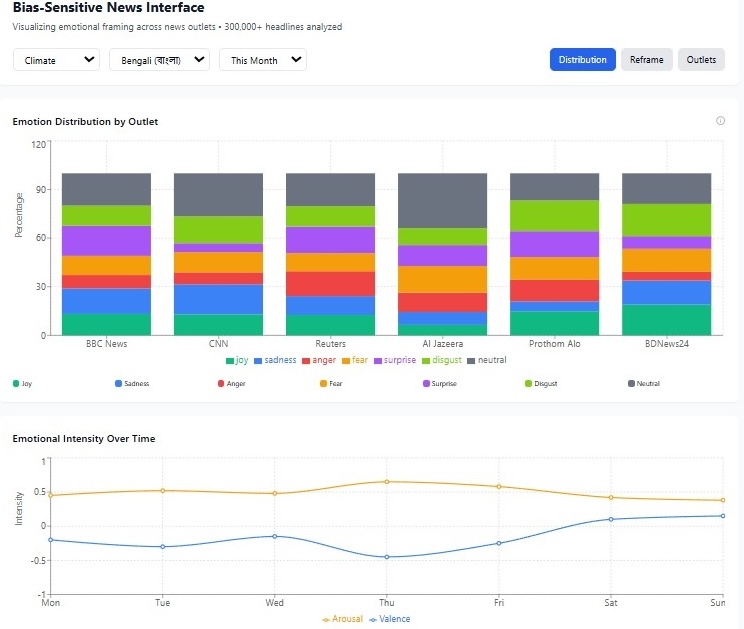}
\caption{Bias Sensitive News Interface showing emotion distribution by outlet and estimated emotional intensity over time.}
\label{fig:bias_interface}
\end{figure}

The \textit{Bias Sensitive News Interface} in Fig.~\ref{fig:bias_interface} provides a broader view of affective patterns across the media ecosystem. It includes outlet level emotion distributions, temporal trends in valence and arousal, and simple divergence based indicators for comparing sources. In this study, the Affective Polarization Index is used as a summary indicator of emotional difference across outlets, while Jensen Shannon Divergence is used to compare outlet level emotion distributions. These indicators are intended to support exploration rather than provide a final judgment about bias.

\subsection{Detailed Emotion Analysis Panel}

\begin{figure}[!t]
\centering
\includegraphics[width=\linewidth]{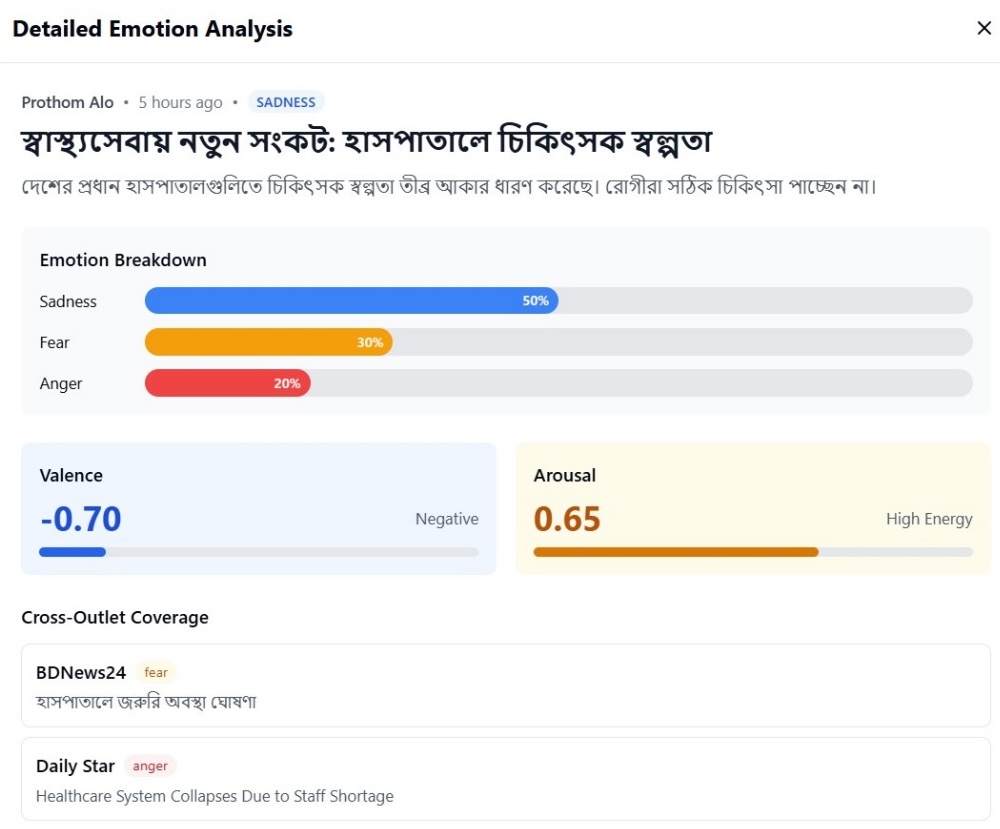}
\caption{Detailed emotion analysis panel showing the estimated emotion distribution for an individual headline and a comparison with related coverage from other outlets.}
\label{fig:emo_detail}
\end{figure}

The \textit{Detailed Emotion Analysis Panel} in Fig.~\ref{fig:emo_detail} focuses on a single headline. It shows the estimated emotion distribution, valence, and arousal values for that headline. It also allows comparison with related coverage from other outlets when similar stories are available. This view may help readers see whether a story is framed with stronger fear, sadness, anger, or neutrality across different sources.

\subsection{Limitations}

This study has several limitations. The analysis relies on zero shot inference with Gemma 3 4B and was checked only on a manually reviewed pilot subset of 200 headlines. A larger annotated Bengali headline dataset would be needed for stronger validation. The study also does not compare Gemma 3 4B with baseline models such as mBERT, XLM R, BanglaBERT, or other Bengali specific models, so the results should not be interpreted as evidence that Gemma 3 4B is the best model for this task.

The analysis reports descriptive statistics only and does not include significance testing, confidence intervals, or robustness analysis. Therefore, the findings should be interpreted as corpus level patterns rather than statistically confirmed claims about the broader Bengali news ecosystem. Since the study focuses only on headlines, the results should also be interpreted as headline level affective patterns, not complete article level bias.

The proposed interface is conceptual and was not evaluated through a user study. As a result, this work cannot claim that the interface improves news literacy, reduces distress, or changes reader behavior. The full dataset cannot be publicly released because it was collected from third party news outlets. Future work may release anonymized metadata, prompts, and processing scripts where permitted. This study also analyzes 300{,}000 headlines from a larger collected corpus of 824{,}784 articles due to computational and time constraints.

\section{Conclusion}\label{sec:conclusion}

This study analyzed 300{,}000 Bengali news headlines using Gemma 3 4B to examine dominant emotion patterns in digital journalism. The results show that negative emotion labels, especially \textit{anger}, \textit{sadness}, \textit{disappointment}, and \textit{fear}, appear frequently in the analyzed corpus. These findings should be interpreted as model based estimates of headline level affective framing rather than evidence of editorial intent.

The study also proposed a conceptual bias sensitive news interface for visualizing emotional framing across sources. Since the interface was not evaluated through a user study, its practical effectiveness remains a topic for future research. Future work will focus on stronger validation, baseline comparison, broader temporal coverage, and user evaluation.

\bibliographystyle{IEEEtran}
\bibliography{references}

\end{document}